\pdfoutput=1
\documentclass[11pt]{article}
\usepackage{acl}
\usepackage{times}
\usepackage{latexsym}
\usepackage{tabularx}
\usepackage{booktabs}
\usepackage{paralist}
\usepackage{todonotes}
\usepackage{tikz}
\usepackage{xspace}
\usetikzlibrary{positioning}
\newcommand{\corpusname}{\mbox{\textsc{DeFaBel}}\xspace}
\newcommand{\F}{F$_1$\xspace}
\usepackage[T1]{fontenc}
\usepackage[utf8]{inputenc}
\usepackage{microtype}
\usepackage{inconsolata}
\usepackage{graphicx}
\usepackage{multirow}
\usepackage{enumitem}
\usepackage{scalefnt}

\title{How Entangled is Factuality and Deception in German? }

\author{%
  Aswathy Velutharambath$^{1,2,3}$,   Amelie Wührl$^{1,3}$ \and Roman Klinger$^{3}$\\
  $^1$Institut f\"ur Maschinelle Sprachverarbeitung, University of
  Stuttgart, Germany \\
  $^2$Psychological AI (100 Worte Sprachanalyse GmbH), Heilbronn, Germany\\
  $^{3}$Fundamentals of Natural Language Processing, University of Bamberg, Germany \\
  \texttt{aswathy.velutharambath@100worte.de},
  \texttt{amelie.wuehrl@ims.uni-stuttgart.de}\\
  \texttt{roman.klinger@uni-bamberg.de}\\
}

\begin{document}
\maketitle
\begin{abstract}
  The statement ``\textit{The earth is flat}'' is factually
  inaccurate, but if someone truly believes and argues in its favor,
  it is not deceptive. Research on deception detection and fact
  checking often conflates factual accuracy with the truthfulness of
  statements. This assumption makes it difficult to (a) study subtle
  distinctions and interactions between the two and (b) gauge their
  effects on downstream tasks.  The belief-based deception framework
  disentangles these properties by defining texts as deceptive when
  there is a mismatch between what people say and what they truly
  believe. In this study, we assess if presumed patterns of deception
  generalize to German language texts. We test the effectiveness of
  computational models in detecting deception using an established
  corpus of belief-based argumentation. Finally, we gauge the impact
  of deception on the downstream task of fact checking and explore if
  this property confounds verification models. Surprisingly, our
  analysis finds no correlation with established cues of
  deception. Previous work claimed that computational models can
  outperform humans in deception detection accuracy, however, our
  experiments show that both traditional and state-of-the-art models
  struggle with the task, performing no better than random
  guessing. For fact checking, we find that natural language
  inference-based verification performs worse on non-factual and
  deceptive content, while prompting large language models for the
  same task is less sensitive to these properties.

\end{abstract}

\section{Introduction}
\begin{figure}
  \centering
  \includegraphics[width=1\linewidth]{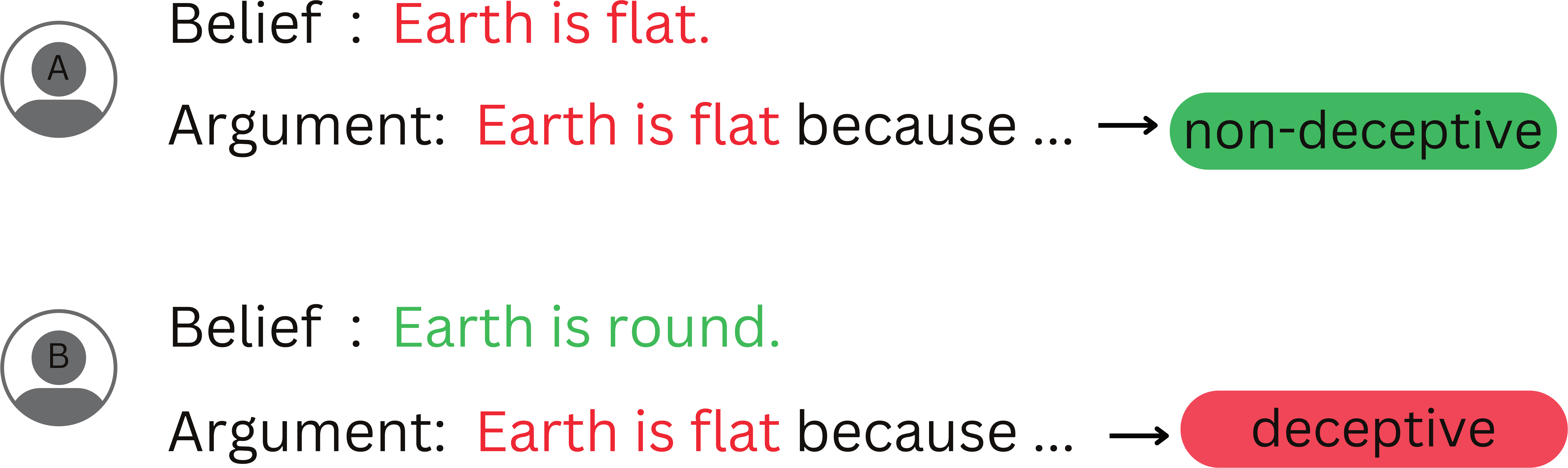}
  \caption{Belief-based deception framework assigns deception label to arguments based on the alignment of the person's belief and argumentation, irrespective of the factual accuracy (green = factually accurate, red = not fact. accurate) of the claim.}
  \label{fig:belief-deception}
\end{figure}
In NLP, a vast body of the existing research on deception has focused
on detecting the veracity of a statement in a given context or domain
\citep{ott-etal-2011-finding, salvetti-etal-2016-tangled,
  perez-rosas-mihalcea-2014-cross,capuozzo-etal-2020-decop}, often
ignoring the difference between factual accuracy and truthfulness,
implicitly equating the two. For example, consider the statement
``\textit{The Colosseum in Rome is an iconic ancient amphitheater. My
  family owns a secret underground passage that leads directly to
  it}''. The first part of the statement is factually
accurate\footnote{\textit{Factual accuracy} here refers to the
  correctness of a statement based on objective evidence that can be
  verified. It indicates that the statement aligns with the actual
  state of affairs and can be proven true or false using reliable
  sources.} and verifiable, while the second part is likely
fabricated, illustrating the importance of disentangling truthfulness
from factual accuracy to accurately assess deceptive intent.
 
Similarly, in fact verification, mis- and disinformation -- differing
in terms of the underlying deceptive intent -- are typically conflated
\citep{boland2022}.  Both in deception detection and fact checking,
this conventional focus leaves a gap in understanding how deceptive
intent operates independently of factual accuracy, which is crucial
for developing more sophisticated models that can discern subtle cues
of deceptions even when embedded within truthful contexts.

In \citet{velutharambath-etal-2024-factual-statements}, we disentangle
factuality and deception by introducing a framework in which deception
is defined as arguing against one's own beliefs (see
Figure~\ref{fig:belief-deception}). This approach emphasizes the
deceptive intent rather than the factual correctness of the statement,
thereby providing a better understanding of how deception can occur
even when parts of a statement are factually accurate.

The belief-based deception corpus (DeFaBel) developed using this
framework provides the opportunity to explore the relationship between
personal beliefs, factual accuracy, and deceptive intent. The dataset
comprises of both deceptive and non-deceptive arguments supporting the
same topic statement. Moreover, the \corpusname corpus is the only
publicly available dataset on deception in German, facilitating an
investigation into the generalizability of linguistic cues across
languages. With this paper, we perform the previously lacking
experiments to assess the effectiveness of automatic deception models
within this corpus.

We make the following contributions: (1) We investigate how linguistic
cues of deception manifest in German-language texts. Surprisingly, we
find no statistically significant correlations between deception
labels and established cues for deceptive content. (2) We assess the
effectiveness of transformer-based and large language models in
identifying deceptive arguments. Our observations indicate that these
models may not be accurately capturing deception cues from language,
as their performance is close to random guessing. (3) Finally, we
explore how evidence-based fact checking is confounded by deception
and factuality. Our findings suggest that non-factual and deceptive
content poses particularly challenging for fact checking
models\footnote{Code \& data:
  \url{https://www.ims.uni-stuttgart.de/data/defabel}}.

\section{Background}

\subsection{Deception}

Deception in communication involves intentionally causing another to
adopt a belief known by the deceiver to be false
\citep{zuckerman-definition1981, mahon2007definition}. This
encompasses lies, exaggerations, omissions, and distortions as various
manifestations of deceptive acts \cite{turner-75, metts-89}. Despite
variations in definition across disciplines, the consensus underscores
the deliberate nature of deception \citep{mahon2007definition,
  gupta2012telling}.

\paragraph{Corpora.} Automatic deception detection from text heavily
relies on labeled corpora. Unlike other NLP tasks, the gold labels
cannot be assigned post-data collection, as the veracity of the
statement hinges on the intention of the author. Deceptive instances
have been solicited via crowd-sourcing, as fake reviews
\citep{ott-etal-2011-finding, ott-etal-2013-negative,
  salvetti-etal-2016-tangled} or false opinions on controversial
topics
\citep{perez-rosas-mihalcea-2014-cross,capuozzo-etal-2020-decop, mu3d}
and by monitoring fake review generation tasks
\citep{yao-etal-2017-online} or users exhibiting suspicious activity
\citep{fornaciari2020fake}. Deceptive instances have also been
extracted from dialogue in strategic deception games such as
Mafiascum\footnote{\url{https://www.mafiascum.net/}}, Box of Lies,
Diplomacy and, To Tell the Truth \citep{Ruiter2018TheMD,
  soldner-etal-2019-box, peskov-etal-2020-takes, Skalicky2020PleasePJ,
  hazra-majumder-2024-tell} based on specific game rules. None of the
mentioned datasets explicitly address the distinction between factual
inaccuracy and deceptive intent. In contrast,
\citet{velutharambath-etal-2024-factual-statements} tackles this by
creating a corpus of argumentative texts, where the deception label is
assigned based on the author's true beliefs, irrespective of factual
accuracy, thereby disentangling the two.

Most deception corpora focus on English, with a comprehensive overview
provided by \citet{velutharambath-klinger-2023-unidecor}. Attempts at
deception detection in other languages have been made, albeit to a
lesser extent, including Polish \citep{polish-deception}, Bulgarian
\citep{bulgariandeception}, Italian \citep{capuozzo-etal-2020-decop},
Russian \cite{pisarevskaya-etal-2017-deception}, Dutch
\citep{verhoeven-daelemans-2014-clips}, and Spanish
\cite{almela-etal-2012-seeing} texts. In this study we make use of the
\corpusname corpus introduced by
\citet{velutharambath-etal-2024-factual-statements}, the only publicly
available corpus to study deception in German.

\paragraph{Linguistic Cues.}
Previous research has explored linguistic cues of deception across
various modalities, including written statements, spoken dialogues,
and online discourse \citep{Newman2003, Bond2005, Zhou2004}. These
cues have proven valuable in automated deception detection, especially
in computer-mediated communication \citep{Zhou2004}. For instance,
\citet{Newman2003} found that reduced self-references in deceptive
statements suggest liars create distance from their falsehoods, while
the use of exclusive words (e.g., \textit{but}, \textit{rather})
introduces ambiguity. \citet{digital-deception-hancock} noted
variations in word count, pronoun usage, emotion word frequency, and
cognitive complexity across discourse types and mediums (e.g.,
real-world vs.\ online, monologue vs.\ dialogue). Linguistic analyses
have been integrated into machine learning models alongside other
features like n-grams \citep{fornaciari-poesio-2014-identifying,
  fornaciari2020fake, ott-etal-2011-finding}, part-of-speech tags
\citep{mu3d, fornaciari2020fake,
  perez-rosas-mihalcea-2015-experiments}, LIWC psychological
categories \citep{perez-rosas-mihalcea-2014-cross,
  yao-etal-2017-online}, and syntactic production rules
\citep{yao-etal-2017-online, perez-rosas-mihalcea-2015-experiments}.
\citet{velutharambath-klinger-2023-unidecor} evaluates the
generalizability of different linguistic cues across multiple
deception datasets. Extensive surveys of linguistic deception cues
include works by \citet{duran_hall_mccarthy_mcnamara_2010},
\citet{pinocchio-2012}, and \citet{linguistic-cues-meta-2015}. In the
current study, we include the most commonly discussed linguistic cues
in our analysis.

\paragraph{Automatic Detection Methods.}
Several studies have explored automated deception detection in textual
data. Some employed feature-based classification methods with
linguistic cues, using support vector machines
\citep{ott-etal-2011-finding, perez-rosas-mihalcea-2014-cross,
  fornaciari-poesio-2014-identifying}, logistic regression
\citep{Ruiter2018TheMD}, decision trees
\citep{perez-rosas-mihalcea-2015-experiments}, and random forests
\citep{soldner-etal-2019-box,
  perez-rosas-mihalcea-2015-experiments}. Others integrated contextual
information with recurrent neural networks
\citep{peskov-etal-2020-takes} and transformer-based models
\citep{capuozzo-etal-2020-decop, peskov-etal-2020-takes,
  fornaciari-etal-2021-bertective}. Transformers are not uniformly
superior; BERT performs comparably to LSTMs
\citep{peskov-etal-2020-takes}, however adding extra attention layers
can enhance performance
\citep{fornaciari-etal-2021-bertective}. Cross-corpus and
within-corpus experiments with the RoBERTa model reveal limited
generalizability across domains
\citep{velutharambath-klinger-2023-unidecor}. Recent studies have
utilized Large Language Models (LLMs) like FLAN-T5 on English-language
datasets, achieving state-of-the-art results, especially with larger
models \citep{loconte2023verbal}. \citet{hazra-majumder-2024-tell}
used zero-shot prompting to extract various cues of deception from
text and a discriminator to aggregate the final prediction. In our
study, we assess feature-based models, fine-tuned transformer models,
and state-of-the-art LLMs for deception detection.

\subsection{Factuality}
Fact verification is concerned with determining if a statement is
factual, i.e., ``generally accepted to be true''
\citep{boland2022}. While some work proposed to predict veracity or
properties such as untrustworthiness based on claim characteristics,
hypothesizing that mis- and disinformation is encoded in the
linguistic properties of a claim \citep{wang-2017-liar,
  rashkin-etal-2017-truth}, the verification process is typically
evidence-based
\citep{guo-etal-2022-survey,vladika-matthes-2023-scientific,
  hardalov-etal-2022-survey}. The input to a fact checking model is a
claim-evidence pair for which we predict a verdict indicating if the
evidence entails, i.e., supports, or contradicts the
claim. Importantly, fact verification usually does not differentiate
intentionally spreading false information (disinformation), e.g., with
a deceptive intent, from other misinformation.

Some work has studied the impact of claim and evidence properties on
fact checking \citep{atanasova-etal-2022-fact,
  kelk-etal-2022-automatic, hansen-etal-2021-automatic} and the effect
of synthetic disinformation on fact checking performance
\cite{du2022synthetic}. However, the effect of a claim’s factual
accuracy and any deceptive intent in the evidence documents remains
unstudied presumably because of a lack of resources that combines
factuality and deception. We leverage the belief-based deception
framework \citep{velutharambath-etal-2024-factual-statements} to
understand the role of deceptive evidence and factuality in the
verification process.

\subsection{Belief-based Deception Framework}\label{sec:dataset-description}

\paragraph{Concept.} In belief-based deception framework
\citep{velutharambath-etal-2024-factual-statements}, deception is
defined as arguing against one's own beliefs, emphasizing the
deceptive intent rather than the factual correctness of the
statement. In Figure~\ref{fig:belief-deception}, the statement
``\textit{The earth is flat}'' is factually inaccurate, but if someone
truly believes and argues in its favor, it is not deceptive. In this
framework, content is deceptive if there is a mismatch between what
people say and what they truly believe.

\paragraph{Dataset description.} In this study, we use the \corpusname
corpus of belief-based deception
\citep{velutharambath-etal-2024-factual-statements}. It is a publicly
available\footnote{The dataset is available under a Creative Commons
  license CC BY-SA. We use the data in its original form without any
  modification.} corpus contains argumentative German texts collected
via crowdsourcing.  Participants were solicited to write persuasive
arguments supporting a given statement irrespective of its factuality
or their personal beliefs. When the argumentation is in contradiction
with their own belief, the instance is labeled as deceptive. The
belief of the annotator is collected after generating the text. Out of
the 30 statements used for soliciting arguments, 19 were factual
(e.g., ``\textit{Camels store fat in their hump.}'') and 11 were
non-factual (e.g., ``\textit{Eating watermelon seeds can cause
  indigestion.}''). The corpus contains 1031 German texts labeled with
deceptive intent. The distribution of labels is skewed, with a higher
frequency of deceptive instances ($\approx$ 62\%) compared to
non-deceptive ones. Also, $\approx$ 60\% of the arguments support
non-factual statements. We employ this corpus for all experiments in
this paper.

\section{How Do Linguistic Cues of Deception Manifest in German Texts? (RQ1)}
\label{sec:linguisit_correlation}
This study evaluates the reliability of linguistic cues widely
presumed to encode deception, for developing an automated deception
detection system in German. As a groundwork to investigate this, we
conduct a correlation analysis to understand the relationship between
each of the linguistic cues and the deception label.

\subsection{Methods}
We consider the most commonly used linguistic cues of deception from
prior studies, as follows.

\paragraph{Complexity.} Previous studies argue that deceptive
statements tend to be simpler compared to truthful ones, attributing
it to the increased cognitive load when deceiving someone which
hinders creativity and complex sentence production
\citep{depaulo2003cues, linguistic-cues-meta-2015}. This implies that
deceptive responses may exhibit reduced lexical diversity, are shorter
or less elaborate relative to truthful responses. To represent the
complexity of arguments, we use token count, sentence count, unique
token count, type-token ratio (lexical diversity) and average POS-tag
counts. As a measure of sophistication of language, we use
Flesch--Kincaid \cite{kincaid1975derivation} and Gunning--Fog
\cite{robert1952technique} readability scores.

\paragraph{Concreteness.}
Linguistic concreteness has been deemed to be important in detecting
deception \cite{kleinberg2019detecting}, but its role is inconclusive
from previous studies. Some argue that liars speak abstractly due to
difficulty in recalling the details, while truth-tellers provide more
specific information \cite{kleinberg2018automated}. The contrary
argues that deceptive narratives may include concrete actions to
reduce the cognitive load \cite{Newman2003}.  To assign a concreteness
score for each argumentative text, we use a German lexical resource
created by \citet{koper-schulte-im-walde-2016-automatically}, which
assigns a concreteness value to a lemmatized word. In addition, we
include the imageability score from the same resource, as concrete
words are often said to have higher imageability compared to abstract
words.

\paragraph{Sentiment.} Studies like \citet{linguistic-cues-meta-2015}
and \citet{vrij2008detecting} have reported that negative emotions are
more prevalent in deceptive speech, possibly induced by guilt and fear
of getting caught. As the arguments in the \corpusname corpus support
emotionally neutral statements, rather than emotions, we investigate
sentiment as a cue to understand if the general sentiment varies in
the case of deceptive and non-deceptive statements. We use a sentiment
classifier to assign sentiment scores to each argumentative
text. Additionally, we include the valence (pleasantness) and arousal
(emotional intensity) scores from
\citet{koper-schulte-im-walde-2016-automatically} as they are closely
related to emotions.

\paragraph{LIWC features.} As noted by \citet{Newman2003}, word usage
patterns differ between truth-tellers and liars. Previous studies have
extensively used LIWC \cite{pennebaker-2015-linguistic}, a general
lexicon capturing different psychological categories, to study these
patterns in deceptive language \citep{linguistic-cues-meta-2015,
  perez-rosas-mihalcea-2014-cross, yao-etal-2017-online}.  Along with
other deception specific cues, we use all psychological categories
available in LIWC, specifically for German, to account for potential
variations in language use between deceptive and non-deceptive
arguments.

\subsection{Experimental Setup}
We use the measure of \textit{point-biserial correlation}
\citep{glass1996statistical} to study the correlation between the
deception label (a discrete value) and the scores assigned to
different features (a continuous value). The point-biserial
correlation $\rho_{\footnotesize\textrm{pb}}$ is
\begin{equation}
  \rho_{\textrm{pb}} = \frac{\mu_{\footnotesize\textrm{decept}} - \mu_{\footnotesize\textrm{truth}}}{\sigma_{n}}\sqrt{\frac{n_{\footnotesize\textrm{decept}} n_{\footnotesize\textrm{truth}}}{n (n-1)}}\,
\end{equation}
where $n$ is the total of instances, $\mu_{\footnotesize\textrm{truth}}$ and
$\mu_{\footnotesize\textrm{decept}}$ are the mean values of the
continuous variable for deceptive and truthful instances respectively,
$\sigma_{n}$ the standard deviation of the continuous variable, and
$n_{\footnotesize\textrm{truth}}$ and
$n_{\footnotesize\textrm{decept}}$ the frequencies of the promotion
and prevention labels, respectively, within the dataset. The
point-biserial correlation coefficient, ranging from $-1$ to $+1$,
signifies perfect negative and perfect positive correlations,
respectively. A high positive correlation coefficient implies that the
score tends to be higher when the instance label is
deceptive. Conversely, a high negative correlation coefficient
suggests that the score is higher when the label is non-deceptive. The
magnitude and sign of the correlation coefficient offer insights into
the strength and direction of the relationship between the deception
label and the cues.

We use the point-biserial correlation implementation from
\texttt{scipy}\footnote{\url{https://docs.scipy.org/doc/scipy-0.14.0/reference/generated/scipy.stats.pointbiserialr.html}}
to analyze the relationship between 128 linguistic cues and the
deception label. We set the desired overall significance level
($\alpha$) to $0.05$ for \textit{Bonferroni correction}
\cite{bonferroni1936teoria}.  Table~\ref{tab:linguistic_cues} in the
Appendix shows details of all linguistic cues.

\subsection{Results}
Surprisingly, from our analysis, we observe that none of the 128 cues
show any statistically significant correlation with the deception
label, implying the absence of a discernible linear relationship
between the cues and deception. Given that \textit{Bonferroni
  corrections} are typically stringent, we opted to consider the
non-adjusted $\alpha$ (.05), resulting in 13 variables with
significant but weak correlation and the scores ranging from $-$.095
to .068. While this may suggest there is a lack of strong association
between linguistic cues and deception, it is essential to note that the
absence of significant correlations does not necessarily imply the
absence of meaningful relationships.

The \corpusname corpus is based on neutral topics that lack strong
emotional commitment, with the stakes of lying simulated by offering
participants incentives to persuade someone else. This approach may
result in deception cues that diverge from those observed in real-life
lying scenarios. This underscores the necessity to explore contextual,
cultural and individualistic factors alongside linguistic cues.

\section{How Efficient are Computational Models in Detecting Deception? (RQ2)}
\label{sec:deception-detection-experiments}
In Section~\ref{sec:linguisit_correlation}, we demonstrate that
previously reported deception cues do not significantly correlate with
the deception labels in the \corpusname dataset. However, previous
studies have shown that automatic methods have achieved some success
in predicting deception. Therefore, we evaluate the performance of
traditional feature-based methods, fine-tuned transformer models, and
instruction-tuned large language models on the \corpusname corpus.

\begin{table}
  \centering\small
  \begin{tabular}{lccc}
    \toprule
    \textbf{Data Split} & \textbf{Deceptive} & \textbf{Non-Deceptive} & \textbf{total} \\
    \cmidrule(lr){1-4}
    Dev. Data           & 491                & 263                    & 754            \\
    Holdout Data        & 152                & 125                    & 277            \\
    \cmidrule(lr){1-4}
    Total               & 643                & 388                    & 1031           \\
    \bottomrule
  \end{tabular}
  \caption{Data split of the \corpusname dataset used for experimentation.}
  \label{tab:datasplit}
\end{table}

\begin{table*}
  \centering \small
\renewcommand{\arraystretch}{1.08}
  \begin{tabular}{lccccccc}
    \toprule
    & \multicolumn{3}{c}{Non-Deceptive} & \multicolumn{3}{c}{Deceptive} & \\
    \cmidrule(r){1-1}    \cmidrule(r){2-4}     \cmidrule(lr){5-7}
    Model                                            & Precision      & Recall       & F-score      & Precision     & Recall       & F-score       & Accuracy     \\
    \cmidrule(r){1-1}    \cmidrule(rl){2-2}\cmidrule(rl){3-3}\cmidrule(rl){4-4}\cmidrule(rl){5-5}
    \cmidrule(rl){6-6}\cmidrule(rl){7-7}\cmidrule(l){8-8}
    random                                           & \textbf{.40}   & .52          & .45          & .64           & .52          & .58           & .52          \\
    majority class                                   & .00            & .00          & .00          & .65           & \textbf{1.00}          & \textbf{.79} & \textbf{.65} \\
    Log.\ Reg                                        & .38            & .29          & .33          & .66           & .75          & .70           & .59          \\
    SVM                                              & .38            & .29          & .33          & .66           & .75          & .70           & .59          \\
    GBERT                                            & \textbf{1.00 } & .00          & .01          & .65           & \textbf{1.00}        & \textbf{.79 }          & \textbf{.65} \\
    Mistral (\texttt{DECEPT})                        & .39            & .50          & .44          & .\textbf{68}  & .58          & .63           & .55          \\
    Mistral (\texttt{DECEPT\_FACT})                  & .37            & .56          & .45          & .\textbf{68 } & .50          & .57           & .52          \\
    Mistral (\texttt{CONV\_DECEPT})                  & .33            & .61          & .43          & .62           & .34          & .44           & .44          \\
    Mistral (\texttt{CONV\_DECEPT\_FACT})            & .35            & \textbf{.88} & \textbf{.50} & .65           & .11          & .19           & .38          \\
    Mistral (\texttt{CONV\_DECEPT\_FACT\_RETHINK})   & .37            & .47          & .42          & .67           & .57          & .62           & .54          \\
    \midrule
    & \multicolumn{6}{c}{Evaluated best model on holdout-data} & \\
    \midrule
    random                                           & .44            & .50          & .47          & .54           & .48          & .50           & .51          \\
    Mistral (\texttt{DECEPT})                        & \textbf{.46}   & .45          & .45          & .55           & .56          & .56           & .51          \\
    \bottomrule
  \end{tabular}
  \caption{Results on evaluating models on development and holdout data.}
  \label{tab:results_edd}
\end{table*}

\subsection{Experimental Setup}
We conduct the deception detection experiment on the \corpusname
corpus, containing 1031 argumentative texts. As shown in Table
\ref{tab:datasplit}, we split the data into a development set
comprising arguments related to 22 statements (754 arguments), and a
holdout set consisting of arguments associated with 8 statements (277
arguments). The topics in the two splits do not have any overlap.

For models that we train or fine-tune, we perform 10-fold
cross-validation. For instruction-tuned models, we evaluate them on
the entire development set. To ensure a fair comparison between
models, we aggregate the predictions from the 10 folds and use the
micro-average across the complete development set. Finally, we
evaluate the best-performing model on the holdout dataset

\paragraph{Models.} We use a logistic regression classifier
(\textbf{Log.\ Reg}) as implemented in
\texttt{scikit-learn}\footnote{\url{https://scikit-learn.org/stable/modules/generated/sklearn.linear_model.LogisticRegression.html}},
with default parameters, to weigh all linguistic cues. For support
vector classification (\textbf{SVM}), we use
\texttt{scikit-learn}\footnote{\url{https://scikit-learn.org/stable/modules/generated/sklearn.svm.SVC.html}}. As
the transformer based model, we utilize the pre-trained German BERT
model \texttt{deepset/GBERT-large} (\textbf{GBERT}).  (Appendix
\ref{app:ex_details_deception} shows modeling details.)

For prompt-based deception detection, we utilize \texttt{Mistral
  8x7B}\footnote{\url{https://huggingface.co/mistralai/Mixtral-8x7B-Instruct-v0.1}},
a state-of-the-art open LLM \cite{jiang2024mixtral}. The model
claims superior performance and multilingual support for English,
French, Italian, German, and Spanish.

We evaluate five different prompts in a one-shot setting (refer to
Appendix \ref{app:promptdesign}). In each case, the model is
instructed to predict the deceptiveness of the provided argument. To
determine whether prompting to predict the factuality of the argument
improves prediction accuracy, we also instruct the model to predict
both the factuality and deceptiveness of the given
argument. Furthermore, we conduct experiments with prompts presented
as single instructions, as well as in conversational or dialogue
formats. Finally, we incorporate chain-of-thought prompting within the
dialogue format. We use the following prompts in this study:

\begin{enumerate}[itemsep=0pt,parsep=0pt,topsep=0pt,partopsep=0pt]
\item \texttt{DECEPT}: single instruction prompt, predicts deceptiveness
\item \texttt{DECEPT\_FACT}: single instruction prompt, predicts factuality and deceptiveness
\item \texttt{CONV\_DECEPT}: predicts deceptiveness in a conversational setting
\item \texttt{CONV\_DECEPT\_FACT}: predicts factuality and deceptiveness in a conversational setting
\item \texttt{CONV\_DECEPT\_FACT\_RETHINK}: incorporates chain-of-thought in the conversational setting
\end{enumerate}

\subsection{Results}
Table \ref{tab:results_edd} shows the results of evaluating different
models on the deception detection task.  From the results on the
development data, we see that none of the models perform better than
the majority class prediction ($.65$) in terms of accuracy. However,
the dataset is quite imbalanced and focusing on the \F score for
deceptive instances would offer more insightful observations regarding
their effectiveness in deception detection. GBERT model achieves an \F
score of $0.79$ on the deception label. This result is almost the same
as majority class prediction, as the model labeled all instances as
deceptive in 9 out of 10 folds. We observe that feature-based models
seem to perform comparably well in identifying deceptive instances but
struggle with non-deceptive ones. However, this could be also be an
artifact of the class imbalance.

The results on prompting instruction-tuned LLMs show that the one-shot
single prompt (\texttt{DECEPT}) seems to work slightly better than
random in terms of accuracy ($\Delta$~3pp acc.) and relatively good at
predicting the deceptive instances ($\Delta$~5pp in \F). Unlike the
pre-trained transformer model (GBERT), this is not an artifact of
predicting instances as only deceptive. On prompting to predict both
factuality and deceptiveness (\texttt{DECEPT\_FACT}) the model
performance drops. To evaluate whether a conversational setting
provides better results, the same prompt was presented as a dialogue
between the user and the assistant. The conversational setting in
itself did not seem to be better than the single prompt. However, when
the model is prompted with chain-of-thought setting, the performance
of the model seems to improve in comparison to the single
conversational prompt \texttt{CONV\_DECEPT}($\Delta$~3pp acc.). This
suggests that optimizing the prompt for a deception detection task
could potentially result in better results. Out of all the models,
Mistral with the simple prompt (\texttt{DECEPT}) seems to achieve the
most reliable results on the deception label. To verify if the model
follows instructions to disregard factuality, we modify the prompt to
include reasoning for deception label predictions. However, we observe
inconsistent adherence with this directive by the model (See Appendix
\ref{app:promptdesign}). Nevertheless, evaluating this ``best'' model
on holdout data gives results similar to random prediction once again.

Our study presents contrasting results compared to previous research
that reported over 80\% accuracy using simple n-gram-based classifiers
\citep{ott-etal-2011-finding, ott-2014-linguistic}. We hypothesize
that these models might leverage domain-specific traits rather than
capturing genuine linguistic cues in deceptive text. Further, more
recent studies employing LSTM and pre-trained language models
\cite{fornaciari-etal-2021-bertective,velutharambath-klinger-2023-unidecor}
have shown promising results in deception detection, albeit often
limited to specific domains, challenging the broader applicability of
these models. In contrast, \citet{perez-rosas-mihalcea-2014-cross},
using a topic-neutral dataset similar to the \corpusname corpus,
achieved 60-70\% accuracy by employing LIWC categories as
features. This superior performance could possibly be attributed to
the fact that the dataset contains opinions on controversial issues
which invokes stronger personal involvement and emotional valence
compared to our neutral-topic corpus.
 
\section{Is Fact Verification Confounded by Deception and Factuality? (RQ3)}
\label{sec:factchecking}
\subsection{Methods}
The core task in fact verification is to assess if a claim is
factual~\citep{thorne-vlachos-2018-automated}. This process determines
the relation between the content of a piece of evidence and the
content of a claim. Fact checking is commonly modeled as an entailment
task, where given a claim-evidence pair a model is trained to predict
whether the evidence (premise) supports or refutes the claim. During
this step, we assume the evidence is given, i.e., selected
beforehand. The fact checking model should predict the entailment
label accurately, regardless of underlying claim and evidence
properties. We hypothesize, however, that models may inadvertently
draw on non-propositional cues of deception or the implicit knowledge
they store from pertaining, thereby influencing their
predictions. This opaque entanglement between deception and factuality
may lead to (a) factual claims being more reliably verified compared
to non-factual ones and (b) cause models to perform worse for
instances in which the evidence is corrupted by a deceptive intention.

To investigate the impact of these properties, we compare the
performance of fact checking models for (a) factual vs.\ non-factual
statements and (b) instances with deceptive vs.\ non-deceptive
evidence.  Further, we inspect these properties in particularly
difficult instances to understand if they might be challenging
instances for the models, particularly because they convey non-factual
or deceptive content.

\subsection{Experimental Setup}

\paragraph{Task.}
We frame fact checking as an entailment/natural language inference
(NLI) task. Each instance is a premise-hypothesis pair. The source
statement from \corpusname is the hypothesis, while the argument is
the premise. The models predict whether the claim is \textsc{entailed}
or \textsc{contradicted} by the evidence premise or if there is a
\textsc{neutral} relation between the two. As all arguments in
\corpusname were written to support a statement, the NLI label for all
instances is \textsc{entailment}.

\paragraph{Models.} We experiment with two off-the-shelf models,
varying in architecture and size: \texttt{mDeBERTa}, a RoBERTa-based
medium-sized model, trained for multilingual
NLI\footnote{\url{https://huggingface.co/MoritzLaurer/mDeBERTa-v3-base-mnli-xnli}}
and \texttt{Mistral7B-Instruct}, an instruction-tuned LLM we use for
few-shot
prompting\footnote{\url{https://huggingface.co/mistralai/Mistral-7B-Instruct-v0.1}}. We
provide all details for the experimental setup as well as prompt
design in Appendix~\ref{app:ex_details_fact_verification}.
To understand the proficiency of these models for our use-case, we test their performance on \corpusname. Table~\ref{tab:fc-performance-separated-by-factual-deceptive} presents the results. Both models show robust performance when evaluated on \corpusname (full) with an \F-score of .85 for mDeBerta and .86\F for the Mistral model on the target class \textsc{entailment}\footnote{We provide the full reports in Table~\ref{tab:fc-performance-full-reports}.}.

\subsection{Results}

\paragraph{How do deception and factuality impact prediction
  performance?} To understand how factuality and deception potentially
confound the verification process, we compare the model performance on
arguments supporting factual statements to their performance for
non-factual statements. Analogously, we compare the performance for
instances with deceptive evidence to instances with non-deceptive
evidence.

Table~\ref{tab:fc-performance-separated-by-factual-deceptive} shows
the results. To remove the effects of the imbalance of
\hbox{(non-)factual} and \hbox{(non-)deceptive} instances in
\corpusname, we report results across all available instances and
subsets of 50 claim-evidence pairs. See
Appendix~\ref{app:ex_details_fact_verification} for the sampling
details.  For mDeBERTa, factual instances are substantially more
reliably verified compared to non-factual instances
($\Delta$~7pp\F). For Mistral, the performance is very similar across
factual and non-factual instances ($\Delta$~1pp\F), with non-factual
instances being slightly better verified.  Regarding the deception
property, both models show a slightly better performance for instances
with non-deceptive evidence ($\Delta$~1pp\F for mDeBERTa and Mistral,
respectively).
When comparing the performance on the subsets, we observe similar trends. Notably though, the Mistral model performs better on the instances with deceptive evidence.
Based on the comparison of our results, we hypothesize that smaller models may be more susceptible to these confounding properties in the verification process, particularly for factuality.

\begin{table}
  \centering\small
  \begin{tabularx}{\linewidth}{Xllllll}
    \toprule
    sample & \multicolumn{3}{c}{mDeBERTa} & \multicolumn{3}{c}{Mistral7B-Instruct} \\
    \cmidrule(lr){1-1}\cmidrule(lr){2-4} \cmidrule(lr){5-7}
      & \multicolumn{1}{c}{P} & \multicolumn{1}{c}{R} & \multicolumn{1}{c}{\F} & \multicolumn{1}{c}{P} & \multicolumn{1}{c}{R} & \multicolumn{1}{c}{\F} \\
    \cmidrule(lr){2-2} \cmidrule(lr){3-3} \cmidrule(lr){4-4} \cmidrule(lr){5-5} \cmidrule(lr){6-6} \cmidrule(lr){7-7}
    full
      & 1.00 & 0.74 & 0.85 &
                             1.00 & 0.76 & 0.86
    \\

    \cmidrule(lr){1-7}
    $+$fact &
    1.00 & 0.81 & 0.90 &
    1.00 & 0.76 & 0.86
    \\
    $-$fact &
    1.00 & 0.70 & 0.83 &
    1.00 & 0.76 & 0.87
    \\

    \cmidrule(lr){1-1}\cmidrule(lr){2-4} \cmidrule(lr){5-7}
    $+$fact$^{50}$ &
              1.00 & 0.80 & 0.89 &
                                   1.00 & 0.72 & 0.84
    \\
    $-$fact$^{50}$ &
              1.00 & 0.72 & 0.84 &
                                   1.00 & 0.74 & 0.85
    \\

    \cmidrule(lr){1-1}\cmidrule(lr){2-4} \cmidrule(lr){5-7}
    $+$decep &
    1.00 & 0.74 & 0.85 &
    1.00 & 0.75 & 0.86
    \\
    $-$decep &
    1.00 & 0.75 & 0.86 &
    1.00&0.77&0.87
    \\

    \cmidrule(lr){1-1}\cmidrule(lr){2-4} \cmidrule(lr){5-7}
    $+$decep$^{50}$ &
               1.00 & 0.68 & 0.81 &
                                    1.00 & 0.78 & 0.88
    \\
    $-$decep$^{50}$ &
               1.00 & 0.74 & 0.85 &
                                    1.00 & 0.76 & 0.86
    \\

    \bottomrule
  \end{tabularx}
  \caption{Verification performance of \texttt{mDeBERTa} and \texttt{Mistral7B-Instruct}. We report results for the full \corpusname dataset (full), for factual/non-factual ($\pm$fact), deceptive/non-deceptive ($\pm$decep) instances  and subsets (50 instances) for each category.}
  \label{tab:fc-performance-separated-by-factual-deceptive}
\end{table}

\paragraph{What are the properties of particularly difficult
  instances?} Finally, we want to understand if instances might get
consistently mis-classified because they convey
non-factual or deceptive content. We inspect the set of
claim-premise pairs that get incorrectly classified by both models
(120 instances).
We find that the majority of instances are non-factual (72~\%) and
deceptive (62~\%). While this is in line with the label distribution
of deceptive instances in \corpusname (62~\%), the percentage of
non-factual instances in the misclassified set is substantially higher
(60~\% in \corpusname). This further corroborates that this property
is a potential error source in the verification process.

We further investigate if errors in the prediction are in fact
correlated with factuality and deception properties of the instances
(see Appendix~\ref{app:ex_details_fact_verification}), but do not find
any meaningful connections.

\section{Conclusion \& Future Work}
\label{sec:conclusion}

Belief-based deception, which disentangles factuality from deceptive
intent, presents the possibility to investigate deceptive intent in
isolation and gauge its impact on downstream applications, such as
automatic fact verification.

In our correlation analysis, we found no clear links between
linguistic cues of deception and expected patterns in German texts,
highlighting the importance of considering cultural and
language-specific differences in deception. Future research should
investigate whether German texts exhibit unique linguistic patterns in
deceptive contexts due to these factors. Furthermore, investigating
belief-based data in the English language -- currently unavailable to
the best of our knowledge -- could offer valuable insights into
whether our negative findings are influenced by the framework. It is
also important to note that the dataset contains labels for instances
at the textual level rather than at the sentence level. Further
investigation is necessary to explore potential variability in the
deceptiveness of individual arguments.

Our automatic deception detection experiments show that computational
models are not yet reliable. Further studies focusing on
explainability using the current dataset could help understand the
linguistic patterns the models are relying on for predicting
deceptiveness. To enhance the robustness of deception detection tools,
future work should prioritize incorporating context, cultural factors,
and individual differences.

To gauge the impact of deception on fact checking, we explore if this
property confounds verification models. A RoBERTa-based model shows
lower performance for instances with non-factual claims and deceptive
evidence documents, indicating that these instances are more difficult
to verify. When using LLM prompting, performance is not substantially
impacted by these properties. However, non-factual instances are a
frequent error class for both models indicating that these properties
might be challenging for the models. Future work should investigate
which additional properties (non-)deceptive evidence or (non-)factual
claims may exhibit to understand if the performance may be impacted
e.g., by the choice of topic, argument structure or persuasive
language.

\section*{Acknowledgments}
This research has been conducted as part of the \textsc{Fibiss}
project which is funded by the German Research Foundation (DFG,
project number: KL 2869/5- 1 -- 438135827). It was further supported
by the \textsc{Ceat} project (DFG, KL~2869/1-2 -- 380093645). 
We also acknowledge the use of the computing resources provided by the 
\mbox{\textit{bwUniCluster}} funded by the Ministry of Science, Research and the 
Arts Baden-Württemberg and the Universities of the State of 
Baden-Württemberg, Germany, within the framework program bwHPC.

\section*{Limitations}
Our study is limited to investigating previously reported linguistic
cues of deception. While we consider it important to include
extra-linguistic factors such as socio-cultural factors, and
individual differences, it is not addressed in this study. Previous
work studied mostly English, meaning cues could be exclusive to this
language and transferable to German only in a limited way.  For fact
verification, we focus on the connection of factuality and deception
with the fact checking label. These properties may also be tied to
topics, which we did not investigate in this study. As outlined in
Section~\ref{sec:conclusion} we see this as an opportunity for future
work. Additionally, we do not include fake news as a form of deception
in this study, as it represents a broader phenomenon involving the
intentional spread of misinformation, distinct from individual acts of
lying.

\section*{Ethical Considerations}
Understanding deception and factuality from both linguistic and
computational perspectives is vital for combating misinformation. Our
work can therefore contribute to more robust and reliable efforts in
detecting deceptive content and counter-acting the spread of false
information. However, we are aware that the same insights could be
misused -- for example, to create more convincing disinformation or
unfairly profile individuals. We therefore emphasize the need for
responsible use of these models.  Developing a system that detects
lies solely based on textual content raises important questions about
its feasibility and ethics. Since lying is not inherently a criminal
act, labeling someone as a liar based on text analysis requires
careful consideration of the implications.

\bibliography{anthology, custom}

\begin{thebibliography}{61}
\providecommand{\natexlab}[1]{#1}

\bibitem[{Akiba et~al.(2019)Akiba, Sano, Yanase, Ohta, and
  Koyama}]{akiba2019optuna}
Takuya Akiba, Shotaro Sano, Toshihiko Yanase, Takeru Ohta, and Masanori Koyama.
  2019.
\newblock \href {https://doi.org/10.1145/3292500.3330701} {{O}ptuna: A
  next-generation hyperparameter optimization framework}.
\newblock In \emph{The 25th ACM SIGKDD International Conference on Knowledge
  Discovery \& Data Mining}, pages 2623--2631.

\bibitem[{Almela et~al.(2012)Almela, Valencia-Garc{\'\i}a, and
  Cantos}]{almela-etal-2012-seeing}
{\'A}ngela Almela, Rafael Valencia-Garc{\'\i}a, and Pascual Cantos. 2012.
\newblock \href {https://aclanthology.org/W12-0403} {Seeing through deception:
  A computational approach to deceit detection in written communication}.
\newblock In \emph{Proceedings of the Workshop on Computational Approaches to
  Deception Detection}, pages 15--22, Avignon, France. Association for
  Computational Linguistics.

\bibitem[{Atanasova et~al.(2022)Atanasova, Simonsen, Lioma, and
  Augenstein}]{atanasova-etal-2022-fact}
Pepa Atanasova, Jakob~Grue Simonsen, Christina Lioma, and Isabelle Augenstein.
  2022.
\newblock \href {https://doi.org/10.1162/tacl_a_00486} {Fact checking with
  insufficient evidence}.
\newblock \emph{Transactions of the Association for Computational Linguistics},
  10:746--763.

\bibitem[{Boland et~al.(2022)Boland, Fafalios, Tchechmedjiev, Dietze, and
  Todorov}]{boland2022}
Katarina Boland, Pavlos Fafalios, Andon Tchechmedjiev, Stefan Dietze, and
  Konstantin Todorov. 2022.
\newblock \href
  {https://www.semantic-web-journal.net/content/beyond-facts-survey-and-conceptualisation-claims-online-discourse-analysis-0}
  {Beyond facts--a survey and conceptualisation of claims in online discourse
  analysis}.
\newblock \emph{Semantic Web -- Interoperability, Usability, Applicability},
  13(5):793--827.

\bibitem[{Bond and Lee(2005)}]{Bond2005}
Gary~D. Bond and Adrienne~Y. Lee. 2005.
\newblock \href {https://doi.org/10.1002/acp.1087} {Language of lies in prison:
  linguistic classification of prisoners{\textquotesingle} truthful and
  deceptive natural language}.
\newblock \emph{Applied Cognitive Psychology}, 19(3):313--329.

\bibitem[{Bonferroni(1936)}]{bonferroni1936teoria}
Carlo Bonferroni. 1936.
\newblock Teoria statistica delle classi e calcolo delle probabilita.
\newblock \emph{Pubblicazioni del R Istituto Superiore di Scienze Economiche e
  Commericiali di Firenze}, 8:3--62.

\bibitem[{Capuozzo et~al.(2020)Capuozzo, Lauriola, Strapparava, Aiolli, and
  Sartori}]{capuozzo-etal-2020-decop}
Pasquale Capuozzo, Ivano Lauriola, Carlo Strapparava, Fabio Aiolli, and
  Giuseppe Sartori. 2020.
\newblock \href {https://aclanthology.org/2020.lrec-1.178} {{D}ec{O}p: A
  multilingual and multi-domain corpus for detecting deception in typed text}.
\newblock In \emph{Proceedings of the Twelfth Language Resources and Evaluation
  Conference}, pages 1423--1430, Marseille, France. European Language Resources
  Association.

\bibitem[{de~Ruiter and Kachergis(2018)}]{Ruiter2018TheMD}
Bob de~Ruiter and George Kachergis. 2018.
\newblock \href {https://arxiv.org/abs/1811.07851} {The mafiascum dataset: A
  large text corpus for deception detection}.
\newblock \emph{ArXiv}, abs/1811.07851.

\bibitem[{DePaulo et~al.(2003)DePaulo, Lindsay, Malone, Muhlenbruck, Charlton,
  and Cooper}]{depaulo2003cues}
Bella~M. DePaulo, James~J. Lindsay, Brian~E. Malone, Laura Muhlenbruck, Kelly
  Charlton, and Harris Cooper. 2003.
\newblock \href {https://doi.org/10.1037/0033-2909.129.1.74} {Cues to
  deception.}
\newblock \emph{Psychological bulletin}, 129(1):74--118.

\bibitem[{Du et~al.(2022)Du, Bosselut, and Manning}]{du2022synthetic}
Yibing Du, Antoine Bosselut, and Christopher~D. Manning. 2022.
\newblock \href {https://doi.org/10.1609/aaai.v36i10.21302} {Synthetic
  disinformation attacks on automated fact verification systems}.
\newblock In \emph{Proceedings of the Thirty-Sixth AAAI Conference on
  Artificial Intelligence}, volume~10.

\bibitem[{Duran et~al.(2010)Duran, Hall, McCarthy, and
  McNamara}]{duran_hall_mccarthy_mcnamara_2010}
Nicholas~D. Duran, Charles Hall, Philip~M. McCarthy, and Danielle~S. McNamara.
  2010.
\newblock \href {https://doi.org/10.1017/S0142716410000068} {The linguistic
  correlates of conversational deception: Comparing natural language processing
  technologies}.
\newblock \emph{Applied Psycholinguistics}, 31(3):439–462.

\bibitem[{Fornaciari et~al.(2021)Fornaciari, Bianchi, Poesio, and
  Hovy}]{fornaciari-etal-2021-bertective}
Tommaso Fornaciari, Federico Bianchi, Massimo Poesio, and Dirk Hovy. 2021.
\newblock \href {https://doi.org/10.18653/v1/2021.eacl-main.232} {{BERT}ective:
  Language models and contextual information for deception detection}.
\newblock In \emph{Proceedings of the 16th Conference of the European Chapter
  of the Association for Computational Linguistics: Main Volume}, pages
  2699--2708, Online. Association for Computational Linguistics.

\bibitem[{Fornaciari et~al.(2020)Fornaciari, Cagnina, Rosso, and
  Poesio}]{fornaciari2020fake}
Tommaso Fornaciari, Leticia Cagnina, Paolo Rosso, and Massimo Poesio. 2020.
\newblock \href {https://doi.org/10.1007/s10579-020-09486-5} {Fake opinion
  detection: how similar are crowdsourced datasets to real data?}
\newblock \emph{Language Resources and Evaluation}, pages 1--40.

\bibitem[{Fornaciari and Poesio(2014)}]{fornaciari-poesio-2014-identifying}
Tommaso Fornaciari and Massimo Poesio. 2014.
\newblock \href {https://doi.org/10.3115/v1/E14-1030} {Identifying fake
  {A}mazon reviews as learning from crowds}.
\newblock In \emph{Proceedings of the 14th Conference of the {E}uropean Chapter
  of the Association for Computational Linguistics}, pages 279--287,
  Gothenburg, Sweden. Association for Computational Linguistics.

\bibitem[{Glass and Hopkins(1996)}]{glass1996statistical}
G.V. Glass and K.D. Hopkins. 1996.
\newblock \href {https://books.google.de/books?id=SFmdAAAAMAAJ}
  {\emph{Statistical Methods in Education and Psychology}}.
\newblock Allyn and Bacon.

\bibitem[{Guhr et~al.(2020)Guhr, Schumann, Bahrmann, and
  Böhme}]{guhr-EtAl:2020:LREC}
Oliver Guhr, Anne-Kathrin Schumann, Frank Bahrmann, and Hans~Joachim Böhme.
  2020.
\newblock \href {https://www.aclweb.org/anthology/2020.lrec-1.202/} {Training a
  broad-coverage german sentiment classification model for dialog systems}.
\newblock In \emph{Proceedings of The 12th Language Resources and Evaluation
  Conference}, pages 1620--1625, Marseille, France. European Language Resources
  Association.

\bibitem[{Guo et~al.(2022)Guo, Schlichtkrull, and
  Vlachos}]{guo-etal-2022-survey}
Zhijiang Guo, Michael Schlichtkrull, and Andreas Vlachos. 2022.
\newblock \href {https://doi.org/10.1162/tacl_a_00454} {A survey on automated
  fact-checking}.
\newblock \emph{Transactions of the Association for Computational Linguistics},
  10:178--206.

\bibitem[{Gupta et~al.(2013)Gupta, Sakamoto, and Ortony}]{gupta2012telling}
Swati Gupta, Kayo Sakamoto, and Andrew Ortony. 2013.
\newblock \href
  {https://users.cs.northwestern.edu/~ortony/Andrew_Ortony_files/2013%20-%20Telling%20it%20like%20it%20isn't.pdf}
  {Telling it like it isn't: A comprehensive approach to analyzing verbal
  deception}.
\newblock Online.

\bibitem[{Hancock(2009)}]{digital-deception-hancock}
Jeffrey~T. Hancock. 2009.
\newblock \href {https://doi.org/10.1093/oxfordhb/9780199561803.013.0019}
  {{Digital deception: Why, when and how people lie online}}.
\newblock In \emph{{Oxford Handbook of Internet Psychology}}. Oxford University
  Press.

\bibitem[{Hansen et~al.(2021)Hansen, Hansen, and
  Chaves~Lima}]{hansen-etal-2021-automatic}
Casper Hansen, Christian Hansen, and Lucas Chaves~Lima. 2021.
\newblock \href {https://doi.org/10.18653/v1/2021.acl-short.12} {Automatic fake
  news detection: Are models learning to reason?}
\newblock In \emph{Proceedings of the 59th Annual Meeting of the Association
  for Computational Linguistics and the 11th International Joint Conference on
  Natural Language Processing (Volume 2: Short Papers)}, pages 80--86, Online.
  Association for Computational Linguistics.

\bibitem[{Hardalov et~al.(2022)Hardalov, Arora, Nakov, and
  Augenstein}]{hardalov-etal-2022-survey}
Momchil Hardalov, Arnav Arora, Preslav Nakov, and Isabelle Augenstein. 2022.
\newblock \href {https://doi.org/10.18653/v1/2022.findings-naacl.94} {A survey
  on stance detection for mis- and disinformation identification}.
\newblock In \emph{Findings of the Association for Computational Linguistics:
  NAACL 2022}, pages 1259--1277, Seattle, United States. Association for
  Computational Linguistics.

\bibitem[{Hauch et~al.(2015)Hauch, Blandón-Gitlin, Masip, and
  Sporer}]{linguistic-cues-meta-2015}
Valerie Hauch, Iris Blandón-Gitlin, Jaume Masip, and Siegfried~L. Sporer.
  2015.
\newblock \href {https://doi.org/10.1177/1088868314556539} {Are computers
  effective lie detectors? a meta-analysis of linguistic cues to deception}.
\newblock \emph{Personality and Social Psychology Review}, 19(4):307--342.

\bibitem[{Hazra and Majumder(2024)}]{hazra-majumder-2024-tell}
Sanchaita Hazra and Bodhisattwa~Prasad Majumder. 2024.
\newblock \href {https://doi.org/10.18653/v1/2024.naacl-long.470} {To tell the
  truth: Language of deception and language models}.
\newblock In \emph{Proceedings of the 2024 Conference of the North American
  Chapter of the Association for Computational Linguistics: Human Language
  Technologies (Volume 1: Long Papers)}, pages 8506--8520, Mexico City, Mexico.
  Association for Computational Linguistics.

\bibitem[{Jiang et~al.(2024)Jiang, Sablayrolles, Roux, Mensch, Savary, Bamford,
  Chaplot, de~las Casas, Hanna, Bressand, Lengyel, Bour, Lample, Lavaud,
  Saulnier, Lachaux, Stock, Subramanian, Yang, Antoniak, Scao, Gervet, Lavril,
  Wang, Lacroix, and Sayed}]{jiang2024mixtral}
Albert~Q. Jiang, Alexandre Sablayrolles, Antoine Roux, Arthur Mensch, Blanche
  Savary, Chris Bamford, Devendra~Singh Chaplot, Diego de~las Casas, Emma~Bou
  Hanna, Florian Bressand, Gianna Lengyel, Guillaume Bour, Guillaume Lample,
  Lélio~Renard Lavaud, Lucile Saulnier, Marie-Anne Lachaux, Pierre Stock,
  Sandeep Subramanian, Sophia Yang, Szymon Antoniak, Teven~Le Scao, Théophile
  Gervet, Thibaut Lavril, Thomas Wang, Timothée Lacroix, and William~El Sayed.
  2024.
\newblock \href {https://arxiv.org/abs/2401.04088} {Mixtral of experts}.
\newblock \emph{Preprint}, arXiv:2401.04088.

\bibitem[{Kelk et~al.(2022)Kelk, Basseri, Lee, Qiu, and
  Tanner}]{kelk-etal-2022-automatic}
Ian Kelk, Benjamin Basseri, Wee Lee, Richard Qiu, and Chris Tanner. 2022.
\newblock \href {https://doi.org/10.18653/v1/2022.fever-1.4} {Automatic fake
  news detection: Are current models {``}fact-checking{''}
  or{``}gut-checking{''}?}
\newblock In \emph{Proceedings of the Fifth Fact Extraction and VERification
  Workshop (FEVER)}, pages 29--36, Dublin, Ireland. Association for
  Computational Linguistics.

\bibitem[{Kincaid et~al.(1975)Kincaid, Fishburne~Jr, Rogers, and
  Chissom}]{kincaid1975derivation}
J.~Peter Kincaid, Robert~P. Fishburne~Jr, Richard~L. Rogers, and Brad~S.
  Chissom. 1975.
\newblock \href {https://stars.library.ucf.edu/istlibrary/56/} {Derivation of
  new readability formulas (automated readability index, fog count and flesch
  reading ease formula) for navy enlisted personnel}.
\newblock Technical Report 8-75, University of Central Florida, Institute for
  Simulation and Training.

\bibitem[{Kleinberg et~al.(2018)Kleinberg, Van Der~Toolen, Vrij, Arntz, and
  Verschuere}]{kleinberg2018automated}
Bennett Kleinberg, Yaloe Van Der~Toolen, Aldert Vrij, Arnoud Arntz, and Bruno
  Verschuere. 2018.
\newblock \href {https://doi.org/10.1002/acp.3407} {Automated verbal
  credibility assessment of intentions: The model statement technique and
  predictive modeling}.
\newblock \emph{Applied cognitive psychology}, 32(3):354--366.

\bibitem[{Kleinberg et~al.(2019)Kleinberg, van~der Vegt, Arntz
  et~al.}]{kleinberg2019detecting}
Bennett Kleinberg, Isabelle van~der Vegt, Arnoud Arntz, et~al. 2019.
\newblock Detecting deceptive communication through linguistic concreteness.
\newblock \emph{PsyArXiv}.

\bibitem[{K{\"o}per and Schulte~im
  Walde(2016)}]{koper-schulte-im-walde-2016-automatically}
Maximilian K{\"o}per and Sabine Schulte~im Walde. 2016.
\newblock \href {https://aclanthology.org/L16-1413} {Automatically generated
  affective norms of abstractness, arousal, imageability and valence for 350
  000 {G}erman lemmas}.
\newblock In \emph{Proceedings of the Tenth International Conference on
  Language Resources and Evaluation ({LREC}'16)}, pages 2595--2598,
  Portoro{\v{z}}, Slovenia. European Language Resources Association (ELRA).

\bibitem[{Lloyd et~al.(2019)Lloyd, Deska, Hugenberg, McConnell, Humphrey, and
  Kunstman}]{mu3d}
Paige~E. Lloyd, Jason~C. Deska, Kurt Hugenberg, Allen~R. McConnell, Brandon~T.
  Humphrey, and Jonathan~W. Kunstman. 2019.
\newblock \href {https://doi.org/10.3758/s13428-018-1061-4} {Miami university
  deception detection database}.
\newblock \emph{Behavior Research Methods}, 51:429--439.

\bibitem[{Loconte et~al.(2023)Loconte, Russo, Capuozzo, Pietrini, and
  Sartori}]{loconte2023verbal}
Riccardo Loconte, Roberto Russo, Pasquale Capuozzo, Pietro Pietrini, and
  Giuseppe Sartori. 2023.
\newblock \href {https://doi.org/10.1038/s41598-023-50214-0} {Verbal lie
  detection using large language models}.
\newblock \emph{Scientific reports}, 13:22849.

\bibitem[{Mahon(2007)}]{mahon2007definition}
James~Edwin Mahon. 2007.
\newblock \href {https://doi.org/10.5840/ijap20072124} {A definition of
  deceiving}.
\newblock \emph{International Journal of Applied Philosophy}, 21(2):181--194.

\bibitem[{Metts(1989)}]{metts-89}
Sandra Metts. 1989.
\newblock \href {https://doi.org/10.1177/026540758900600202} {An exploratory
  investigation of deception in close relationships}.
\newblock \emph{Journal of Social and Personal Relationships}, 6(2):159--179.

\bibitem[{Newman et~al.(2003)Newman, Pennebaker, Berry, and
  Richards}]{Newman2003}
Matthew~L. Newman, James~W. Pennebaker, Diane~S. Berry, and Jane~M. Richards.
  2003.
\newblock \href {https://doi.org/10.1177/0146167203029005010} {Lying words:
  Predicting deception from linguistic styles}.
\newblock \emph{Personality and Social Psychology Bulletin}, 29(5):665--675.

\bibitem[{Ott(2014)}]{ott-2014-linguistic}
Myle Ott. 2014.
\newblock \href {https://doi.org/10.3115/v1/W14-2606} {Linguistic models of
  deceptive opinion spam}.
\newblock In \emph{Proceedings of the 5th Workshop on Computational Approaches
  to Subjectivity, Sentiment and Social Media Analysis}, page~31, Baltimore,
  Maryland. Association for Computational Linguistics.

\bibitem[{Ott et~al.(2013)Ott, Cardie, and Hancock}]{ott-etal-2013-negative}
Myle Ott, Claire Cardie, and Jeffrey~T. Hancock. 2013.
\newblock \href {https://aclanthology.org/N13-1053} {Negative deceptive opinion
  spam}.
\newblock In \emph{Proceedings of the 2013 Conference of the North {A}merican
  Chapter of the Association for Computational Linguistics: Human Language
  Technologies}, pages 497--501, Atlanta, Georgia. Association for
  Computational Linguistics.

\bibitem[{Ott et~al.(2011)Ott, Choi, Cardie, and
  Hancock}]{ott-etal-2011-finding}
Myle Ott, Yejin Choi, Claire Cardie, and Jeffrey~T. Hancock. 2011.
\newblock \href {https://aclanthology.org/P11-1032} {Finding deceptive opinion
  spam by any stretch of the imagination}.
\newblock In \emph{Proceedings of the 49th Annual Meeting of the Association
  for Computational Linguistics: Human Language Technologies}, pages 309--319,
  Portland, Oregon, USA. Association for Computational Linguistics.

\bibitem[{Pennebaker et~al.(2015)Pennebaker, Boyd, Jordan, and
  Blackburn}]{pennebaker-2015-linguistic}
James~W. Pennebaker, Ryan Boyd, Kayla Jordan, and Kate Blackburn. 2015.
\newblock \href {https://doi.org/10.15781/T29G6Z} {\emph{{The development and
  psychometric properties of} LIWC2015}}.
\newblock University of Texas at Austin.

\bibitem[{P{\'e}rez-Rosas and Mihalcea(2014)}]{perez-rosas-mihalcea-2014-cross}
Ver{\'o}nica P{\'e}rez-Rosas and Rada Mihalcea. 2014.
\newblock \href {https://doi.org/10.3115/v1/P14-2072} {Cross-cultural deception
  detection}.
\newblock In \emph{Proceedings of the 52nd Annual Meeting of the Association
  for Computational Linguistics (Volume 2: Short Papers)}, pages 440--445,
  Baltimore, Maryland. Association for Computational Linguistics.

\bibitem[{P{\'e}rez-Rosas and
  Mihalcea(2015)}]{perez-rosas-mihalcea-2015-experiments}
Ver{\'o}nica P{\'e}rez-Rosas and Rada Mihalcea. 2015.
\newblock \href {https://doi.org/10.18653/v1/D15-1133} {Experiments in open
  domain deception detection}.
\newblock In \emph{Proceedings of the 2015 Conference on Empirical Methods in
  Natural Language Processing}, pages 1120--1125, Lisbon, Portugal. Association
  for Computational Linguistics.

\bibitem[{Peskov et~al.(2020)Peskov, Cheng, Elgohary, Barrow,
  Danescu-Niculescu-Mizil, and Boyd-Graber}]{peskov-etal-2020-takes}
Denis Peskov, Benny Cheng, Ahmed Elgohary, Joe Barrow, Cristian
  Danescu-Niculescu-Mizil, and Jordan Boyd-Graber. 2020.
\newblock \href {https://doi.org/10.18653/v1/2020.acl-main.353} {It takes two
  to lie: One to lie, and one to listen}.
\newblock In \emph{Proceedings of the 58th Annual Meeting of the Association
  for Computational Linguistics}, pages 3811--3854, Online. Association for
  Computational Linguistics.

\bibitem[{Pisarevskaya et~al.(2017)Pisarevskaya, Litvinova, and
  Litvinova}]{pisarevskaya-etal-2017-deception}
Dina Pisarevskaya, Tatiana Litvinova, and Olga Litvinova. 2017.
\newblock \href {https://doi.org/10.26615/978-954-452-038-0_001} {Deception
  detection for the {R}ussian language: Lexical and syntactic parameters}.
\newblock In \emph{Proceedings of the 1st Workshop on Natural Language
  Processing and Information Retrieval associated with {RANLP} 2017}, pages
  1--10, Varna, Bulgaria. INCOMA Inc.

\bibitem[{Rashkin et~al.(2017)Rashkin, Choi, Jang, Volkova, and
  Choi}]{rashkin-etal-2017-truth}
Hannah Rashkin, Eunsol Choi, Jin~Yea Jang, Svitlana Volkova, and Yejin Choi.
  2017.
\newblock \href {https://doi.org/10.18653/v1/D17-1317} {Truth of varying
  shades: Analyzing language in fake news and political fact-checking}.
\newblock In \emph{Proceedings of the 2017 Conference on Empirical Methods in
  Natural Language Processing}, pages 2931--2937, Copenhagen, Denmark.
  Association for Computational Linguistics.

\bibitem[{Robert(1968)}]{robert1952technique}
Gunning Robert. 1968.
\newblock \href {https://www.worldcat.org/title/1153113711} {\emph{{The
  Technique of Clear Writing}}}.
\newblock McGraw-Hill, New York.

\bibitem[{Salvetti et~al.(2016)Salvetti, Lowe, and
  Martin}]{salvetti-etal-2016-tangled}
Franco Salvetti, John~B. Lowe, and James~H. Martin. 2016.
\newblock \href {https://aclanthology.org/L16-1558} {A tangled web: The faint
  signals of deception in text - boulder lies and truth corpus ({BLT}-{C})}.
\newblock In \emph{Proceedings of the Tenth International Conference on
  Language Resources and Evaluation ({LREC}'16)}, pages 3510--3517,
  Portoro{\v{z}}, Slovenia. European Language Resources Association (ELRA).

\bibitem[{Sarzynska-Wawer et~al.(2023)Sarzynska-Wawer, Pawlak, Szymanowska,
  Hanusz, and Wawer}]{polish-deception}
Justyna Sarzynska-Wawer, Aleksandra Pawlak, Julia Szymanowska, Krzysztof
  Hanusz, and Aleksander Wawer. 2023.
\newblock \href {https://doi.org/10.1371/journal.pone.0281179} {Truth or lie:
  Exploring the language of deception}.
\newblock \emph{PLOS ONE}, 18(2):1--17.

\bibitem[{Skalicky et~al.(2020)Skalicky, Duran, and
  Crossley}]{Skalicky2020PleasePJ}
Stephen~Cameron Skalicky, Nicholas~D. Duran, and Scott~Andrew Crossley. 2020.
\newblock \href {https://doi.org/10.5210/dad.2020.205} {Please, please, just
  tell me: The linguistic features of humorous deception}.
\newblock \emph{Dialogue Discourse}, 11:128--149.

\bibitem[{Soldner et~al.(2019)Soldner, P{\'e}rez-Rosas, and
  Mihalcea}]{soldner-etal-2019-box}
Felix Soldner, Ver{\'o}nica P{\'e}rez-Rosas, and Rada Mihalcea. 2019.
\newblock \href {https://doi.org/10.18653/v1/N19-1175} {Box of lies: Multimodal
  deception detection in dialogues}.
\newblock In \emph{Proceedings of the 2019 Conference of the North {A}merican
  Chapter of the Association for Computational Linguistics: Human Language
  Technologies, Volume 1 (Long and Short Papers)}, pages 1768--1777,
  Minneapolis, Minnesota. Association for Computational Linguistics.

\bibitem[{Swol et~al.(2012)Swol, Braun, and Malhotra}]{pinocchio-2012}
Lyn M.~Van Swol, Michael~T. Braun, and Deepak Malhotra. 2012.
\newblock \href {https://doi.org/10.1080/0163853X.2011.633331} {Evidence for
  the pinocchio effect: Linguistic differences between lies, deception by
  omissions, and truths}.
\newblock \emph{Discourse Processes}, 49(2):79--106.

\bibitem[{Temnikova et~al.(2023)Temnikova, Gargova, Margova, Kireva, Dzhumerov,
  Stefanova, and Krasteva}]{bulgariandeception}
Irina Temnikova, Silvia Gargova, Ruslana Margova, Veneta Kireva, Ivo Dzhumerov,
  Tsvetelina Stefanova, and Hristiana Krasteva. 2023.
\newblock New bulgarian resources for studying deception and detecting
  disinformation.
\newblock In \emph{10th LANGUAGE AND TECHNOLOGY CONFERENCE: Human Language
  Technologies as a Challenge for Computer Science and Linguistics}. Adam
  Mickiewicz University Press.

\bibitem[{Thorne and Vlachos(2018)}]{thorne-vlachos-2018-automated}
James Thorne and Andreas Vlachos. 2018.
\newblock \href {https://aclanthology.org/C18-1283} {Automated fact checking:
  Task formulations, methods and future directions}.
\newblock In \emph{Proceedings of the 27th International Conference on
  Computational Linguistics}, pages 3346--3359, Santa Fe, New Mexico, USA.
  Association for Computational Linguistics.

\bibitem[{Turner et~al.(1975)Turner, Edgley, and Olmstead}]{turner-75}
Ronny~E. Turner, Charles Edgley, and Glen Olmstead. 1975.
\newblock \href {http://www.jstor.org/stable/23255229} {Information control in
  conversations: Honesty is not always the best policy}.
\newblock \emph{The Kansas Journal of Sociology}, 11(1):69--89.

\bibitem[{Velutharambath and
  Klinger(2023)}]{velutharambath-klinger-2023-unidecor}
Aswathy Velutharambath and Roman Klinger. 2023.
\newblock \href {https://doi.org/10.18653/v1/2023.wassa-1.5} {{UNIDECOR}: A
  unified deception corpus for cross-corpus deception detection}.
\newblock In \emph{Proceedings of the 13th Workshop on Computational Approaches
  to Subjectivity, Sentiment, {\&} Social Media Analysis}, pages 39--51,
  Toronto, Canada. Association for Computational Linguistics.

\bibitem[{Velutharambath et~al.(2024)Velutharambath, W{\"u}hrl, and
  Klinger}]{velutharambath-etal-2024-factual-statements}
Aswathy Velutharambath, Amelie W{\"u}hrl, and Roman Klinger. 2024.
\newblock \href {https://aclanthology.org/2024.lrec-main.243} {Can factual
  statements be deceptive? the {D}e{F}a{B}el corpus of belief-based deception}.
\newblock In \emph{Proceedings of the 2024 Joint International Conference on
  Computational Linguistics, Language Resources and Evaluation (LREC-COLING
  2024)}, pages 2708--2723, Torino, Italia. ELRA and ICCL.

\bibitem[{Verhoeven and Daelemans(2014)}]{verhoeven-daelemans-2014-clips}
Ben Verhoeven and Walter Daelemans. 2014.
\newblock \href {http://www.lrec-conf.org/proceedings/lrec2014/pdf/1_Paper.pdf}
  {{CL}i{PS} stylometry investigation ({CSI}) corpus: A {D}utch corpus for the
  detection of age, gender, personality, sentiment and deception in text}.
\newblock In \emph{Proceedings of the Ninth International Conference on
  Language Resources and Evaluation ({LREC}'14)}, pages 3081--3085, Reykjavik,
  Iceland. European Language Resources Association (ELRA).

\bibitem[{Vladika and Matthes(2023)}]{vladika-matthes-2023-scientific}
Juraj Vladika and Florian Matthes. 2023.
\newblock \href {https://doi.org/10.18653/v1/2023.findings-acl.387} {Scientific
  fact-checking: A survey of resources and approaches}.
\newblock In \emph{Findings of the Association for Computational Linguistics:
  ACL 2023}, pages 6215--6230, Toronto, Canada. Association for Computational
  Linguistics.

\bibitem[{Vrij(2008)}]{vrij2008detecting}
Aldert Vrij. 2008.
\newblock \emph{Detecting lies and deceit: Pitfalls and opportunities}.
\newblock John Wiley \& Sons.

\bibitem[{Wang(2017)}]{wang-2017-liar}
William~Yang Wang. 2017.
\newblock \href {https://doi.org/10.18653/v1/P17-2067} {{``}liar, liar pants on
  fire{''}: A new benchmark dataset for fake news detection}.
\newblock In \emph{Proceedings of the 55th Annual Meeting of the Association
  for Computational Linguistics (Volume 2: Short Papers)}, pages 422--426,
  Vancouver, Canada. Association for Computational Linguistics.

\bibitem[{Yao et~al.(2017)Yao, Dai, Huang, and Caverlee}]{yao-etal-2017-online}
Wenlin Yao, Zeyu Dai, Ruihong Huang, and James Caverlee. 2017.
\newblock \href {https://doi.org/10.26615/978-954-452-049-6_102} {Online
  deception detection refueled by real world data collection}.
\newblock In \emph{Proceedings of the International Conference Recent Advances
  in Natural Language Processing, {RANLP} 2017}, pages 793--802, Varna,
  Bulgaria. INCOMA Ltd.

\bibitem[{Zhou et~al.(2004)Zhou, Burgoon, Nunamaker, and Twitchell}]{Zhou2004}
Lina Zhou, Judee~K. Burgoon, Jay~F. Nunamaker, and Doug Twitchell. 2004.
\newblock \href {https://doi.org/10.1023/b:grup.0000011944.62889.6f}
  {Automating linguistics-based cues for detecting deception in text-based
  asynchronous computer-mediated communications}.
\newblock \emph{Group Decision and Negotiation}, 13(1):81--106.

\bibitem[{Zuckerman et~al.(1981)Zuckerman, DePaulo, and
  Rosenthal}]{zuckerman-definition1981}
Miron Zuckerman, Bella~M. DePaulo, and Robert Rosenthal. 1981.
\newblock \href {https://doi.org/10.1016/S0065-2601(08)60369-X} {Verbal and
  nonverbal communication of deception}.
\newblock In Leonard Berkowitz, editor, \emph{Advances in Experimental Social
  Psychology}, volume~14, pages 1--59. Academic Press.

\end{thebibliography}

\clearpage
\appendix

\section{Modeling Details}
\label{app:add_exp_details}

\subsection{Deception Detection}
\label{app:ex_details_deception}

\paragraph{GBERT.} Fine-tuning is conducted using the 
\texttt{BertForSequenceClassification}\footnote{\url{https://huggingface.co/transformers/model_doc/bert.html}} 
implementation from Hugging Face. During fine-tuning, we set the 
number of epochs to 8, the learning rate to $10^{-5}$, and the batch 
size to 16. To prevent overfitting, we monitor the training loss and stop 
if it does not decrease for 5 consecutive batches. We retain default 
values for all other hyperparameters unless specified.

Additionally, we conduct hyperparameter optimization using 
\texttt{Optuna} \citep{akiba2019optuna}. The optimization process 
involves defining a search space for several hyperparameters, including 
learning rate, number of epochs, train batch size, and eval batch 
size. The objective function evaluates different combinations of these 
hyperparameters by training the model and minimizing the training 
loss. Optuna's \texttt{study.optimize} method is employed to run 
multiple trials, automatically selecting the best set of 
hyperparameters based on the lowest training loss observed during 
the trials.

In the main body of the paper, we present the results obtained with the 
default hyperparameters, as both the best and default parameters yielded 
similar results. The model consistently predicted most instances as 
\textit{deceptive}, making the default configuration a pragmatic choice for 
reporting results.

\paragraph{Computational details.} The deception detection experiments 
were conducted on a cluster with distributed memory. The node we used 
featured an Intel Xeon Gold 6230 processor and a NVIDIA Tesla V100 
accelerator. The GBERT model required approximately 30 minutes of GPU 
runtime to complete 10-fold cross-validation and evaluation on the 
holdout set, and approximately 5.5 hours for hyperparameter optimization. 
The prompt-based Mistral model consumed approximately 2.5 hours of GPU 
runtime to evaluate 5 different prompt settings. The prompts 
used for these evaluations are provided as part of the supplementary material.

\subsection{Fact Verification}
\label{app:ex_details_fact_verification}

\paragraph{mDeBERTa.} We use
\texttt{mDeBERTa}\footnote{\url{https://huggingface.co/MoritzLaurer/mDeBERTa-v3-base-mnli-xnli}},
a RoBERTa-based medium-sized model, trained for multilingual NLI. We
use the \texttt{transformers} library and provide the model with
tokenized premise-hypothesis pairs. We convert the model output into
probabilities for each class (\textsc{entailment}, \textsc{neutral},
\textsc{contradiction}) represented by the logits using Softmax. We
run the experiments on a single Nvidia GeForce RTX A6000 GPU.

\paragraph{Mistral.} We use
\texttt{Mistral-7B-instruct}\footnote{\url{https://huggingface.co/mistralai/Mistral-7B-Instruct-v0.1}}
to prompt for NLI labels in the fact verification setting.  In a
conversational one-shot prompt setup, a fictional user describes
setting, task, formatting instructions and an example instance. We
imitate a one-turn conversation in which the LLM assistant provides
the correct answer for the example instance. Subsequently, the user
provides the assistant with the actual instance. Refer to
Table~\ref{tab:promptdetailsfc} for the prompt template. We provide
the initialized prompts as part of the supplementary material. For
each input prompt, we apply the respective chat
template\footnote{\url{https://huggingface.co/docs/transformers/main/en/chat_templating}}
and generate the output sequence with following parameters:
max\_new\_tokens$=$1024, temperature$=$0.3, do\_sample=True,
top\_p$=$0.95, top\_k$=$50, repetition\_penalty$=$1.2.

We instruct the model to provide the output in JSON format. If the
model does not generate a label within the label space, or does not
provide JSON formatted text, we randomly assign the instance to one of
the three target classes. Note that this concerns a total of 31
instances.

We run our experiments on a GPU server with 4 Nvidia GeForce GTX 1080
Ti GPU nodes. Generating the responses for all instances in
\corpusname takes approx. 5 hours.

\paragraph{Subsampling.}
\corpusname consists of an unbalanced number of factual
vs.\ non-factual source statements. To understand if this impacts the
results, we draw a sample to evaluate on. We further want to ensure
that the source statements do not overlap between the subsets. Out of
the 30 source statements (11 factual, 19 non-factual), we sample 10
statements and subsequently sample 5 arguments per statement, leaving
us with 50 pairs each for evaluation.

\paragraph{Are prediction errors correlated with deception \&
  factuality?} We investigate if prediction errors are correlated with
a) factuality of statements and b) deceptive intent in evidence
premises. We calculate Pearson’s correlation between the two binary
variables ((in)correctly predicted and (non-)factual/(non-)deceptive).
For mDeBerta, we observe a correlation of 0.12 with the factuality
property (p-value$<$0.05).  We do not find significant correlations
for the other property and model, indicating that there might be more
complex textual properties indirectly linked to the deception label
that impact the verification process.

\begin{table*}
  \centering\small
  \setlength{\tabcolsep}{10pt}
  \begin{tabular}{lllllrlllr}
    \toprule

     & & \multicolumn{4}{c}{mDeBERTa} & \multicolumn{4}{c}{Mistral7B-Instruct} \\
   \cmidrule(lr){3-6} \cmidrule(l){7-10}
      & & P & R & \F & S & P & R & \F & S \\
    \cmidrule(lr){2-2} \cmidrule(lr){3-3} \cmidrule(lr){4-4} \cmidrule(lr){5-5} \cmidrule(lr){6-6} \cmidrule(lr){7-7}
    \cmidrule(lr){8-8} \cmidrule(lr){9-9} \cmidrule(lr){10-10}


\multirow{6}{*}{\rotatebox[origin=c]{90}{full}} & Neutral &
 0.00 & 0.00 & 0.00 &    0 &
0.00 & 0.00 & 0.00 &    0\\

& Contradict &
0.00 & 0.00 & 0.00 &    0 &
0.00 & 0.00 & 0.00 &    0\\

& Entailed &
1.00 & 0.74 & 0.85 & 1031 &
    1.00 & 0.76 & 0.86 & 1031\\

&
    micro avg &  &  &  0.74 & 1031 &
&  &    0.76 & 1031\\
&   macro avg &  0.33 & 0.25 & 0.28 & 1031
    &  0.33 & 0.25 & 0.29 & 1031\\
& weighted avg &  1.00 & 0.74 & 0.85 & 1031
    &  1.00 & 0.76 & 0.86 & 1031
    \\

    \cmidrule(lr){2-2} \cmidrule(lr){3-6} \cmidrule(l){7-10}

\multirow{6}{*}{\rotatebox[origin=c]{90}{$+$fact}}
    & Neutral & 0.00 & 0.00 & 0.00 &    0 &
0.00 & 0.00 & 0.00 &    0
    \\
& Contradict & 0.00 & 0.00 & 0.00 &    0 &
0.00 & 0.00 & 0.00 &    0
    \\
& Entailed & 1.00 & 0.81 & 0.90 & 376 &
1.00 & 0.76 & 0.86 & 376
    \\
 &   micro avg &  &  &    0.81 &  376
&  &  &    0.76 & 376
    \\
 &  macro avg &  0.33 & 0.27 & 0.30 &  376 &
0.33 & 0.25 & 0.29 & 376
    \\
& weighted avg &  0.33 & 0.27 & 0.30 & 376 &
1.00 & 0.76 & 0.86 & 376
    \\

    \cmidrule(lr){2-2} \cmidrule(lr){3-6} \cmidrule(l){7-10}
\multirow{6}{*}{\rotatebox[origin=c]{90}{$-$fact}} &    Neutral & 0.00 & 0.00 & 0.00 &    0 &
0.00 & 0.00 & 0.00 &    0
    \\
& Contradict & 0.00 & 0.00 & 0.00 &    0 &
0.00 & 0.00 & 0.00 &    0
    \\
& Entailed & 1.00 & 0.70 & 0.83 & 655 &
1.00 & 0.76 & 0.87 & 655
    \\
&     micro avg &  &  &    0.70 &  655
& &  &    0.76 & 655
    \\
&    macro avg &  0.33 & 0.23 & 0.28 & 655
& 0.33 & 0.25 & 0.29 & 655
    \\
& weighted avg &  1.00 & 0.70 & 0.83 & 655
&  1.00 & 0.76 & 0.87 & 655
    \\

\cmidrule(lr){2-2} \cmidrule(lr){3-6} \cmidrule(l){7-10}
\multirow{6}{*}{\rotatebox[origin=c]{90}{$+$decep}}
    &    Neutral & 0.00 & 0.00 & 0.00 &    0 &
0.00 & 0.00 & 0.00 &    0
    \\
& Contradict & 0.00 & 0.00 & 0.00 &    0 &
0.00 & 0.00 & 0.00 &    0
    \\
    & Entailed & 1.00 & 0.74 & 0.85 &  643 &
1.00 & 0.75 & 0.86 &  643
    \\
&     micro avg &  &  &    0.74 &  643 &
&  &    0.75 &  643
    \\
&    macro avg &  0.33 & 0.25 & 0.28 &  643 &
0.33 & 0.25 & 0.29 &  643
    \\
& weighted avg &  1.00 & 0.74 & 0.85 &  643 &
1.00 & 0.75 & 0.86 &  643
    \\

\cmidrule(lr){2-2} \cmidrule(lr){3-6} \cmidrule(l){7-10}
\multirow{6}{*}{\rotatebox[origin=c]{90}{$-$decep}}
    &    Neutral & 0.00 & 0.00 & 0.00 &    0 &
0.00 & 0.00 & 0.00 &    0
    \\
& Contradict & 0.00 & 0.00 & 0.00 &    0 &
0.00 & 0.00 & 0.00 &    0
    \\
    & Entailed & 1.00 & 0.75 & 0.86 &  388 &
1.00 & 0.77 & 0.87 &  388
    \\
&      micro avg &  &  & 0.75 &  388 &
&  &    0.77 &  388
    \\
&    macro avg &  0.33 & 0.25 & 0.29 &  388 &
0.33 & 0.26 & 0.29 &  388
    \\
& weighted avg &  1.00 & 0.75 & 0.86 &  388 &
1.00 & 0.77 & 0.87 &  388
    \\

    \bottomrule
  \end{tabular}
  \caption{Verification performance (\underline{P}recision, \underline{R}ecall, \F, \underline{S}upport) of \texttt{mDeBERTa} and \texttt{Mistral7B-Instruct}. We report results for the full \corpusname dataset (full), for factual/non-factual ($\pm$fact) and deceptive/non-deceptive ($\pm$decep) instances.}
  \label{tab:fc-performance-full-reports}
\end{table*}

\section{Prompt design}
\label{app:promptdesign}
Table~\ref{tab:promptdetailsdeception} shows details on the prompts
for deception detection and \ref{tab:promptdetailsfc} for
fact checking. Please refer to
Sections~\ref{sec:deception-detection-experiments} and
\ref{sec:factchecking} for
explanations how these are used in our experiments.\\[\baselineskip]

\paragraph{Is factuality ignored consistently?}

To make sure that the model follows instructions and ignore the factual accuracy of
statements, we modify the last line of \texttt{DECEPT} prompt to \texttt{\{``Is the text deceptive?''': Yes or No, "Reason": \}}. On manually inspecting the reasons, we see that while the model sticks to the instructions in most cases, it is not consistent in ignoring the factual accuracy. For instance, it uses arguments like the following: 
\begin{compactitem}
	\item ``\textit{the author ignores basic facts about marine life and fish behavior to deceive readers}''
	\item ``\textit{the statement is not supported by factual evidence.}''
	\item ``\textit{the text is factually accurate and includes evidence to support the claim}''
	\item ``\textit{the text presents a clear, factual statement}''
\end{compactitem} 

\begin{table*}
\centering\small\scalefont{0.9}
  \begin{tabularx}{\textwidth}{lX}
    \toprule
    \textbf{Name}               & \textbf{Prompt structure}\\
    \midrule
    \texttt{DECEPT} & \texttt{{`\textbf{system\_prompt}': "You are analyzing German text to identify deceptive texts based on linguistic cues only.",
                      `\textbf{user\_prompt}': ``The task is to evaluate a text and determine whether they exhibit signs of deception. Focus solely on the linguistic properties of the text and disregard any considerations of factual accuracy. Look for patterns of language, logic, or presentation that may indicate an attempt to mislead or manipulate the reader.
                      In each text, pay attention to the cues of deception like:
                      \textbf{<list of cues>}
                      Keep in mind that factual inaccuracy alone cannot be considered a sign of deception. Some individuals may genuinely believe misinformation. Therefore, focus solely on linguistic cues to identify deception. After analyzing each text for these cues, determine whether they exhibit signs of deception or not. Provide your assessment based solely on the linguistic properties of the text, disregarding any considerations of factual accuracy.
                      Look at the following example:
                      \textbf{<example>}
                      This text is deceptive because of the following reason:
                      \textbf{<reasons>}
                      Now judge the following text referring to the information above:
                      ``\textbf{<argument>}''.
                      Judge the presence of different linguistic cues of deception. Think also whether the arguments are supporting a factual statement. Make sure factual accuracy of the statement is not affecting your judgement about the deceptive nature of the text.
                      Strictly format your answer exactly as the following json and do not include any explanations in the output. \{``Is the text deceptive?''': Yes or No\}''
                      }}
    \\
    \texttt{DECEPT\_FACT}       & \texttt{DECEPT}   +     \texttt{\{``Are the arguments in the text factual?'': Yes or No, ``Is the text deceptive? '': Yes or No\}''}         \\
    \texttt{CONV\_DECEPT} & \texttt{{``\textbf{role}'': ``\textbf{user}'',
                            ``\textbf{content}'': 'You are an AI agent who can analyse German text to identify deceptive texts based on linguistic cues only.'},
                            {``\textbf{role}'': ``\textbf{assistant}'',
                            ``\textbf{content}'': 'As an AI analyzing German text, I confidently spot deception using linguistic cues like inconsistencies and evasion. Yet, I also recognize the importance of considering context and evidence for precise evaluation.'},
                            {``\textbf{role}'': ``\textbf{user}'',
                            ``\textbf{content}'': 'Your task is to evaluate a text and determine whether they exhibit signs of deception. Focus solely on the linguistic properties of the text. It is very important that you strictly discard factual accuracy of the statement as there are instances where the author believes in a factually inaccurate statement and is arguing in favor of it, which is non-deceptive. Do you understand? '},
                            {``\textbf{role}'': ``\textbf{assistant}'',
                            ``\textbf{content}'': 'Yes, I understand. I will focus solely on the linguistic properties of the text to evaluate whether it exhibits signs of deception, disregarding the factual accuracy of the statement.'},
                            {``\textbf{role}'': ``\textbf{user}'',
                            ``\textbf{content}'': ``In each text, pay attention to the cues of deception like:
                            \textbf{<list of cues>}
                            Keep in mind that factual inaccuracy alone cannot be considered a sign of deception. Some individuals may genuinely believe misinformation. Therefore, focus solely on linguistic cues to identify deception. After analyzing each text for these cues, determine whether they exhibit signs of deception or not. Provide your assessment based solely on the linguistic properties of the text, disregarding any considerations of factual accuracy.''},
                            {``\textbf{role}'': ``\textbf{assistant}'',
                            ``\textbf{content}'': 'Understood. Please provide the texts for analysis, and I will evaluate them based solely on the linguistic cues you have outlined.'},
                            {``\textbf{role}'': ``\textbf{user}'',
                            ``\textbf{content}'': ``Judge whether or not the following text is deceptive. Return result as a json {{``Is the text deceptive'': **Yes or No**}}
                            text: ``\textbf{<example>}''
                            },
                            {``\textbf{role}'': ``\textbf{assistant}'',
                            ``\textbf{content}'': Answer: {``Is the text deceptive'': ``Yes''}
                            This text is deceptive because of the following reason:
                            \textbf{<reasons>}
                            {``\textbf{role}'': ``\textbf{user}'',
                            ``\textbf{content}'': Judge whether or not the following text is deceptive.
                            text: ``\textbf{<argument>}''
                            Return result as a json \{``Is the text deceptive'': **Yes or No**\}. Keep explanation short.}}} \\
    \texttt{CONV\_DECEPT\_FACT} & \texttt{CONV\_DECEPT} + \texttt{\{``Are the arguments in the text factual?'': Yes or No, ``Is the text deceptive? '': Yes or No\}''} \\
    \texttt{CONV\_DECEPT\_RETHINK} & \texttt{CONV\_DECEPT} + \texttt{                    {``\textbf{role}'': ``\textbf{user}'',
                                     ``\textbf{content}'': ``Are the arguments in the text supporting a factual statement?''},
                                     {``\textbf{role}'': ``\textbf{assistant}'',
                                     ``\textbf{content}'': ``Yes. They are arguing that the earth is round. This statement is factually accurate.''},
                                     {``\textbf{role}'': ``\textbf{user}'',
                                     ``\textbf{content}'': ``Is the text deceptive irrespective of the factuality of the statement it is supporting? Strictly format your answer exactly as the following json and do not include any explanations in the output ``Answer: {{``Is the text deceptive?'': Yes or No}}''},
                                     {``\textbf{role}'': ``\textbf{assistant}'',
                                     ``\textbf{content}'': Answer: {{``Is the text deceptive?'': ``Yes''}}''},
                                     {``\textbf{role}'': ``\textbf{user}'',
                                     ``\textbf{content}'': Here is another text. Judge the presence of different linguistic cues of deception. Think also whether the arguments are supporting a factual statement. Make sure factual accuracy of the statement is not affecting your judgment about the deceptive nature of the text. }} $\ldots$ \\
    \bottomrule
  \end{tabularx}
  \caption{Structure of prompts used for the deception detection experiments.}
  \label{tab:promptdetailsdeception}
\end{table*}

\begin{table*}
  \centering\footnotesize
  \begin{tabularx}{\linewidth}{lX}
    \toprule
    user      & \texttt{You're going to be working on a natural language inference (NLI), also known as textual entailment task. As you probably know that this is a subfield of natural language processing that deals with determining if one sentence or text implies another sentence or text. Your goal is to determine whether a hypothesis statement is entailed or refuted by premise sentence(s). You will be presented with a premise and a hypothesis pair. Note that the texts are in German. Then your task is to categorize pair into one of three categories: 1) Entailment: The hypothesis is entailed by the premise, 2) Neutral: The hypothesis is neutral to the premise, or 3) Contradiction: The hypothesis contradicts the premise. Provide the output in a json format with the key being `label' and the value being the category number as an integer. For example, if you believe the post should be categorized as 3) Contradiction , your json output should be: \{`label': 3\} Now consider the following pair: Premise: <<Die Beweise für die Rundum-Theorie der Erde sind überwältigend. Astronomische Beobachtungen, Raumfahrt und Phänomene der Zeitzonen bestätigen alle die kugelförmige Gestalt der Erde. Bilder von Weltraummissionen zeigen deutlich die Krümmung unseres Planeten und bestätigen damit jahrhundertelangen wissenschaftlichen Konsens. Fringe-Theorien, die eine flache Erde behaupten, stützen sich jedoch auf ausgewählte Daten und emotionale Appelle, anstatt auf empirische Beweise. Obwohl es wichtig ist, alternative Standpunkte zu berücksichtigen, widerspricht die Leugnung der Realität einer runden Erde etabliertem wissenschaftlichem Wissen.>>Hypothesis: <<Die Erde ist eine Kugel.>> What is the correct label for this pair?} \\
    \cmidrule(lr){1-1}\cmidrule(lr){2-2}
    assistant & \texttt{Here is the correct answer formatted as json: \{`label': 1\}}     \\
    \cmidrule(lr){1-1}\cmidrule(lr){2-2}
    user & \texttt{Perfect! Here is another post: Premise: \textbf{<premise>} Hypothesis: \textbf{<hypothesis>}What is the correct label for this pair? Only output the json, no additional text or explanation.}
    \\
    \bottomrule
  \end{tabularx}
  \caption{Prompt template for generating NLI-based fact verification labels. We provide the instantiated prompts in the supplementary material.}
  \label{tab:promptdetailsfc}
\end{table*}

\section{Linguistic cues}
\label{sec:app_cues}
We show details on the operationalization of the linguistic cues in Table~\ref{tab:linguistic_cues}. Please see Section~\ref{sec:linguisit_correlation} for details on the use of these features.\\[\baselineskip]

\begin{table*}
  \centering\footnotesize
  \begin{tabularx}{\linewidth}{lll}
    \toprule
    \textbf{Linguistic cues} & \textbf{Features}                    & \textbf{Operationalization}                                  \\
    \cmidrule{1-3}
    Complexity               & Token count                          & \# tokens                                                    \\
                             & Sentence count                       & \# sentences                                                 \\
                             & Unique token count                   & \# unique tokens                                             \\
                             & Type-token ratio / lexical diversity & \# unique tokens / \# tokens                                 \\
                             & Average postag count                 & \# postag / \# tokens calculated per postag                  \\
                             & Flesch--Kincaid readability           & calculated using \texttt{textstat} python library            \\
                             & Gunning--Fox readability              & calculated using \texttt{textstat} python library            \\
    \cmidrule{1-3}
    Concreteness             & Abstractness score                    & from \citet{koper-schulte-im-walde-2016-automatically}       \\

                             & Imageability                         & from \citet{koper-schulte-im-walde-2016-automatically}       \\

    \cmidrule{1-3}
    Sentiment                & Sentiment score                      & Positive, negative, and neutral scores using \citet{guhr-EtAl:2020:LREC} \\
                             & Arousal                              & from \citet{koper-schulte-im-walde-2016-automatically}       \\
                             & Valence                              & from \citet{koper-schulte-im-walde-2016-automatically}       \\
    \cmidrule{1-3}
    LIWC                     & 99 psychological categories          & Relative frequency for from DE-LIWC2015                      \\
    \bottomrule
  \end{tabularx}
  \caption{Linguistic cues and their operationalization}
  \label{tab:linguistic_cues}
\end{table*}

\end{document}